\pdfoutput=1

\documentclass[11pt]{article}

\usepackage[final]{EMNLP2022}

\usepackage{times}
\usepackage{latexsym}

\usepackage{algorithm}
\usepackage{algorithmic}
\usepackage{hyperref}
\usepackage{adjustbox}

\usepackage[T1]{fontenc}

\usepackage[utf8]{inputenc}

\usepackage{microtype}

\usepackage{inconsolata}

\makeatletter
\newenvironment{myframework}[1][htb]{%
    \renewcommand{\ALG@name}{Framework}
  \begin{algorithm}[#1]%
  }{\end{algorithm}}
  \makeatother

\title{EncT5: A Framework for Fine-tuning T5 as Non-autoregressive Models}

\author{Frederick Liu, Terry Huang, Shihang Lyu, Siamak Shakeri, Hongkun Yu, Jing Li \\
        Google  \\ \{frederickliu, hterry, shihang, siamaks, hongkuny, jingli\}@google.com}

\begin{document}
\maketitle
\begin{abstract}

Pre-trained encoder-decoder transformer architectures have become increasingly popular recently with the advent of T5 models. T5 has also become more favorable over other architectures like BERT due to the amount of data that it is pre-trained on, increased scale of model parameter sizes and easy applicability to a diverse set of tasks due to the generative nature of the model. While being able to generalize to a wide variety of tasks, it is not clear that encoder-decoder architectures are the most efficient for fine-tuning tasks that don't require auto-regressive decoding. In this work, we study fine-tuning pre-trained encoder-decoder models for tasks such as classification, multi-label classification, and structured prediction. We propose \textbf{EncT5}, a framework for these problems, and illustrate instantiations for these tasks. Our experiment results show that EncT5 has advantages over T5 such as efficiency and usability out performs BERT when evaluated on publicly available pre-trained checkpoints. 
\end{abstract}




\section{Introduction}

The scaling law of Transformers \cite{vaswani2017attention}, availablity of massive textual corpora (such as C4~\cite{2020t5} or mC4 \cite{xue-etal-2021-mt5}) and increased compute available from accelerators \cite{kaplan2020scaling} to perform unsupervised pre-training have contributed to the steady progress and adoption of pretrained langauge models in NLP. Software frameworks like Mesh Tensorflow \cite{shazeer2018mesh} helped unlock greater model parallelism and pushed even further the limits of training large language model.

Pre-trained models \cite{devlin-etal-2019-bert, yang2019xlnet, clark2020electra} reduce the dependence on large amounts of task-specific training data and instead only need a more modest fine-tuning dataset. 
Because training large models from scratch can be data and compute heavy, users prefer to consume and adapt existing pretrained-models for their desired task.
Platforms such as TF Hub\footnote{https://www.tensorflow.org/hub} and HuggingFace \cite{wolf-etal-2020-transformers} provide various families of such pre-trained models, and have become increasingly popular.


\begin{figure}[!t]
\centering
\includegraphics[width= 0.8\linewidth]{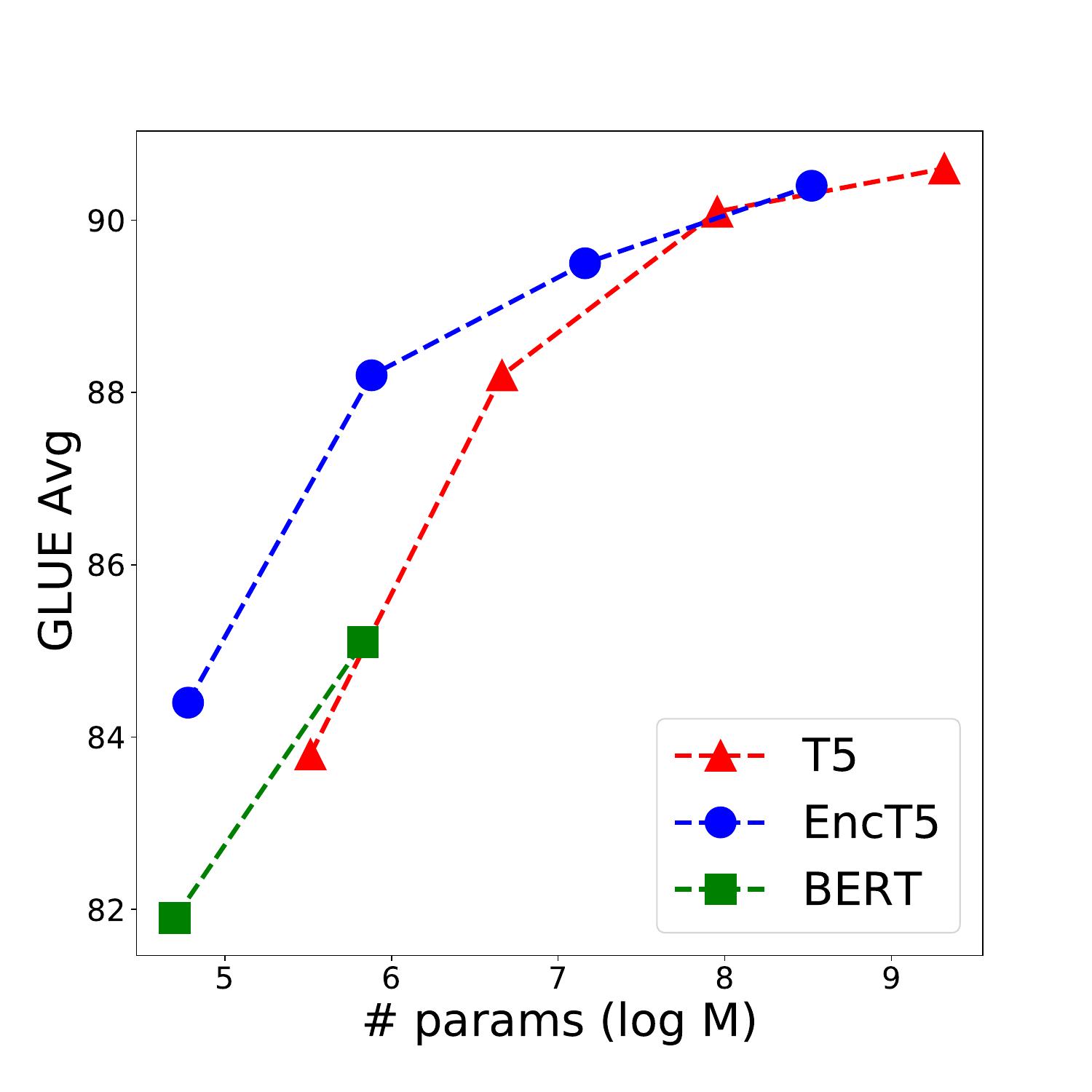}
   \caption{Number of parameters and GLUE average for EncT5, T5 and BERT at various scales.}
   \label{fig:params_glue} 
\end{figure}

\begin{figure*}[!t]
   \includegraphics[width= 1.0\linewidth]{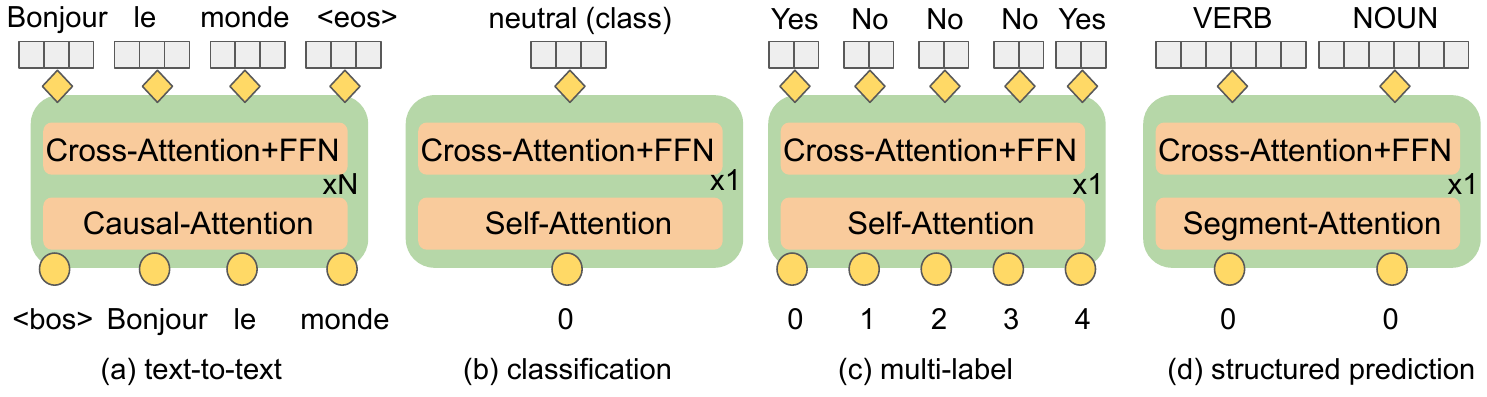}
   \caption{Instantiations of the \textbf{decoder} in \textbf{EncT5} framework (showing its inputs, outputs and internal composition) and comparison to T5. The token ids for (b)(c)(d) no longer have explicit semantic meaning and it represents the latent vector it maps to. The encoder is omitted since there is no difference and the output of the encoder is passed to Cross-Attention block.}
   \label{fig:decoders} 
\end{figure*}

T5 \cite{2020t5} introduced a unified framework for converting a wide variety of NLP tasks (such as generation, regression and classification ones) into a text-to-text prediction task. While previous encoder-only models such as BERT \cite{devlin-etal-2019-bert} were only able to solve non-generation problems, T5's unified framework permitted its model to train through a much more diverse set of tasks.

Despite the general flexibility of the text-to-text problem setup, we find in our experiments that this setup is an unnatural fit for discriminative tasks (such as classification, regression, and structured prediction tasks). We find that the decoder layers of the proposed encoder-decoder architecture of T5, are under-utilized in these types of tasks. 

In this work, we study how to harvest T5's pre-trained model parameters and produce a model optimized for fine-tuning of discriminative tasks. The resulting model uses fewer parameters (thus resulting in better training and serving efficiency) and has minimal quality loss (in some tasks quality improvements). We propose \textbf{EncT5}, a framework which reuses T5 encoder layers with non-intrusive code change. Our proposed approach preserves the pre-training of the encoder-decoder model, and can easily be applied to all T5 variants such as mT5 \cite{xue-etal-2021-mt5} or ByT5 \cite{xue2021byt5}.

Our contributions are:
\begin{itemize}
    \item Propose a framework to convert T5 models to be used in a non-autoregressive fashion with the same Transformer interface such that T5-style models can be used to solve problems in more task-appropriate ways with "non-intrusive code changes". 
    \item We show that EncT5 is a better in fit for classification, multi-label, and structured prediction tasks than T5 in quality, usability, and efficiency.
\end{itemize}


\section{Text-to-Text Transfer Transformer}\label{sec:t2t}

T5 is an encoder-decoder Transformer pre-trained on the Colossal Clean Crawled Corpus (C4) dataset with span-corruption objective. One important benefit of such encoder-decoder architecture Figure~\ref{fig:decoders} (a) is that it can be applied to generative tasks, such as summarization, as well as discriminative tasks such as natural language inference.

Because the T5 model's interface expects input and outputs to be text, natural language inference tasks are converted to a text-to-text format. For example, the target label in a classification task and the score in a regression task is cast to a string and later tokenized. The tokenization transformation can sometimes produce illegal labels (e.g. the labels are not represented in the vocabulary)~\cite{xue-etal-2021-mt5} and it is also not always clear how to convert multi-label or structured prediction tasks into a target sequence. We will discuss the challenges of applying T5 on these tasks in future sections.

Below we give an introduction of the encoder-decoder interface and how it is used in general.

\subsection{Encoder-Decoder Transformer Interface}
T5 follows the Encoder-Decoder Transformer interface introduced by ~\cite{vaswani2017attention} by taking the following inputs:  
\begin{itemize}
    \item \texttt{encoder\_input\_tokens}: input data to the encoder. In general, it is the tokenized input text.
    \item \texttt{decoder\_input\_tokens}: input token to the decoder. During training, it is the tokenized target text shifted to the right and pre-pended with a begin-of-sentence token. During inference, it is the tokenized text that the model has generated so far. 
    \item \texttt{decoder\_target\_tokens}: target text tokenized. Used to calculate loss during training.
    \item \texttt{encoder\_segment\_ids}: Generally used to provide segmentation info for packed examples but can also be used to control how \texttt{encoder\_input\_tokens} can attend to each other.
    \item \texttt{decoder\_segment\_ids}: Generally used to provide decoder segmentation info for packed examples but can also be used to control how \texttt{decoder\_input\_tokens} attend to each other and how \texttt{decoder\_input\_tokens} attends to \texttt{encoder\_input\_tokens}.
    \item \texttt{encoder\_positions}: Encoder subsequence positions for packed examples. Can be inferred from  \texttt{encoder\_input\_tokens} when packing is not enabled.
    \item \texttt{decoder\_positions}: Decoder subsequence positions for packed examples. Can be inferred from  \texttt{decoder\_input\_tokens} when packing is not enabled.
    
\end{itemize}

\section{EncT5 Framework}\label{sec:enct5}

In this section, we propose a framework (Framework 1) for converting a T5 checkpoint into a EncT5 checkpoint for non-autoregressive tasks and show instantiations for single-label classification/regression (Algorithm ~\ref{alg:cls}), multi-label classification (Algorithm ~\ref{alg:mlabel}), and structured prediction tasks (Algorithm ~\ref{alg:tag}). We also describe in more detail the modifications to the model inputs used by EncT5 for each of the above-mentioned tasks. 

EncT5 uses the same Encoder-Decoder interface as T5 and supplies almost the same inputs to the model.  Because T5 is autoregressive and EncT5 is not, the inputs in EncT5 are tailored for the non-autoregressive task setup.

In T5 at each timestep, \texttt{decoder\_input\_tokens} maps the input token fed to the decoder and \texttt{decoder\_target\_tokens} maps to the output token to be predicted.

In EncT5, we were inspired by recent work in feeding latent vectors to Transformers as queries \cite{carion2020end, pmlr-v139-jaegle21a} and have modified the decoder inputs to be latent vectors that learn how to best pool encoded information through self-attention and cross-attention.

\texttt{decoder\_input\_tokens} are modified to be the latent vector id (out of a total latent vector vocab size) selected at each timestep. After self- and cross-attention are applied to the latent vectors, a projection to the final class label size is then applied.  

Alternatives to the EncT5 framework are to simply reuse the encoder and either: (1) pre-pend a \texttt{[CLS]} token, mimicking BERT in the beginning of the input; or (2) max or mean pool the encoder tokens. For (1), the encoder does not see it during pre-training and this would cause a discrepancy between pre-training and fine-tuning. For (2), it has been shown that basic pooling is sub-optimal for sentence representation~\cite{reimers2019sentence}. Both alternatives also require an intrusive change and are unfriendly to packing, a trick frequently used to improve accelerator utilization by packing multiple examples into a single example. We instead opted for a simpler design which minimized complex code changes and avoided the above mentioned drawbacks by decoupling the encoder logic and the pooling logic.

\begin{myframework}[t] 
	\caption{Fine-tune T5 with EncT5}
	\label{alg:main-framework}
	\begin{algorithmic}[1]
		\REQUIRE Pre-trained checkpoint from T5 family and inputs for the Encoder Decoder Transformer interface.
		\STATE Select number of decoders.
		\STATE Initiate decoder embedding from scratch and define the corresponding latent vector vocabulary size.
		\STATE Initiate decoder projection head from scratch and define the projection size.
		\STATE Select weights to load from checkpoint.
		\STATE Select loss function and minimize loss between  \texttt{decoder\_target\_tokens} and \texttt{decoder\_output\_logits}.
	\end{algorithmic}
\end{myframework}

\subsection{Classification/Regression}
We describe the instantiation of fine-tuning classification and regression tasks   in Algorithm ~\ref{alg:cls} with EncT5 and the semantic meaning of the Transformer inputs. We find that the decoder is not helpful in this scenario. Our intuition is that, with a single token, the self-attention layer is simply an identity and the depth does not capture extra interaction.\footnote{For efficiency, users can remove the self-attention layer in the decoder.} 

\setcounter{algorithm}{0}
\begin{algorithm}[t] 
	\caption{Fine-tune Classification/Regression}
	\label{alg:cls}
	\begin{algorithmic}[1]
	    \STATE Set number of decoders to 1.
	    \STATE Initiate decoder embedding from scratch and set latent vector vocab size to 1.
	    \STATE Initiate decoder projection head from scratch and set projection size to number of classes for classification and 1 for regression.
	    \STATE Load encoder weights.
	    \STATE Use cross-entropy for classification and mean squared error for regression.
	\end{algorithmic}
\end{algorithm}
\begin{itemize}
\item \texttt{encoder\_input\_tokens}: Tokenized input text.
\item \texttt{decoder\_input\_tokens}: A single 0 token, which the mapped embedding will be learned to decide how to pool information from the encoder. 
\item \texttt{decoder\_target\_tokens}: Target label or score.
\item \texttt{encoder\_segment\_ids}: Used for packing.
\item \texttt{decoder\_segment\_ids}: Used for packing.
\item \texttt{encoder\_positions}: Used for packing.
\item \texttt{decoder\_positions}: None.
\end{itemize}

\subsection{Multi-label Classification}
We describe the instantiation of fine-tuning multi-label classification   in Algorithm ~\ref{alg:mlabel} with EncT5 and the semantic meaning of the Transformer inputs. Unlike Algorithm ~\ref{alg:cls}, the length of the 
\texttt{decoder\_input\_tokens} is greater than 1 and the self-attention in the decoder is no longer an identity. Thus, increasing the number of decoder layers can result in performance gain. On the other hand, the attention complexity in the decoder is $O(n^2)$, where n is \texttt{decoder\_input\_tokens}. This instantiation is a problem when \texttt{decoder\_input\_tokens} is large. To tackle problem with large output space, one can always fallback to Alogirthm ~\ref{alg:mlabel} which does not capture label interactions by setting the projection size to number of labels and loss to binary cross entropy. Another design difference is the projection that each \texttt{decoder\_output\_logits} uses. In the multi-label setup, we want each label to be treated as a binary classification problem so the projection head cannot be shared.   
\setcounter{algorithm}{1}
\begin{algorithm}[t] 
	\caption{Fine-tune Multi-label Classification}
	\label{alg:mlabel}
	\begin{algorithmic}[1]
	    \STATE Set number of decoders to 1.
	    \STATE Initiate decoder embedding from scratch and set latent vector vocab size to number of possible labels.
	    \STATE Initiate decoder projection head from scratch and set projection size to 2. Each \texttt{decoder\_input\_token\_id} should map to its own projection head.
	    \STATE Load encoder weights.
	    \STATE Use cross-entropy and treat each loss as a binary classification problem.
	\end{algorithmic}
\end{algorithm}
\setcounter{algorithm}{2}
\begin{algorithm}[t] 
	\caption{Fine-tune Structured Prediction}
	\label{alg:tag}
	\begin{algorithmic}[1]
	    \STATE Set number of decoders to 1.
	    \STATE Initiate decoder embedding from scratch and set latent vector vocab size to 1.
	    \STATE Initiate decoder projection head from scratch and set projection size to number of classes per text span.
	    \STATE Use cross-entropy.
	\end{algorithmic}
\end{algorithm}

\begin{table*}[!ht]
\caption{Results on the GLUE validation set. For BERT numbers, we follow Section 4.1 of the BERT paper~\cite{devlin-etal-2019-bert} for training setup. Similar to T5, SST-2 is bucketized and treated as a classification task for EncT5 and T5.}
\begin{adjustbox}{width=\textwidth}
\begin{tabular}{l|ccccccccc|c}
               & MNLI-m & MNLI-mm & QQP & QNLI & SST-2 & CoLA & STS-B & MRPC & RTE  & GLUE   \\ 
                            & \multicolumn{2}{c}{Acc} & F1 & Acc & Acc & Mathew  & Spearman & F1  & Acc & Avg  \\ \hline 
            BERT-base & 83.1 &	83.7 &	87.8	& 90.4 &	92.2 &	56.9 &	84.6 &	90.6 &	70.0 &	82.1 \\
            T5-base & 88.8	& 88.4	& 89.5	& 93.2	& 94.7	& 53.4	& 85.2	& 91.6	& 69.7	& 83.8\\
            EncT5-base & 88.6	& 88.7	& 89.6	& 93.4	& 94.6	& 56.6	& 84.4	& 89.7	& 74.4	& \textbf{84.4}\\
            
            \hline
            BERT-large & 87.1	& 86.9	& 88.4	& 93.0	& 94.0	& 63.9	& 86.4	& 92.4	& 75.5	& 85.1\\ 
            T5-large & 91.0	& 90.9	& 89.6& 94.8	& 96.9	& 63.6	& 88.0	& 93.5	& 85.6	& \textbf{88.2}\\
            EncT5-large & 91.1	& 91.4	& 90.0	& 95.4	& 97.2	& 63.6	& 88.0	& 93.3	& 86.3	& 88.5\\
            \hline
            T5-xl & 92.1	& 91.7	& 90.3	& 96.2	& 97.0	& 70.9	& 87.4	& 93.5	& 92.1	& \textbf{90.1}\\
            EncT5-xl & 91.9	& 91.7	& 90.4	& 96.1	& 97.1	& 69.2	& 86.9	& 93.1	& 89.2	& 89.5\\
            \hline
            T5-xxl & 92.1	& 92.0	& 90.4	& 96.4	& 97.2	& 72.9	& 86.9	& 94.2	& 92.8	& \textbf{90.6}\\
            EncT5-xxl & 92.4	& 92.1	& 90.1	& 96.7	& 97.5	& 71.9	& 86.9	& 93.7	& 92.8	& 90.4
            
\end{tabular}
\end{adjustbox}
\label{tab:glue}
\end{table*}

\begin{itemize}
\item \texttt{encoder\_input\_tokens}: Tokenized input text.
\item \texttt{decoder\_input\_tokens}: During training, we feed \textbf{range(} \texttt{num\_labels} \textbf{)} and label ids we are interested in during inference.
\item \texttt{decoder\_target\_tokens}: A boolean label for each \texttt{decoder\_input\_tokens}.
\item \texttt{encoder\_segment\_ids}: Used for packing.
\item \texttt{decoder\_segment\_ids}: Used for packing.
\item \texttt{encoder\_positions}: Used for packing.
\item \texttt{decoder\_positions}: None.
\end{itemize}
\subsection{Structured Prediction (Part Of Speech Tagging)}
We demonstrate how to design an instantiation for structured prediction using the Part of speech (POS) task as an example. The design should, however, be transferable to other structured prediction tasks. The main difference between Algorithm~\ref{alg:tag} with Algorithm~\ref{alg:cls} is that instead of predicting the whole input sentence, we predict a label for each text span. In order to constraint the \texttt{decoder\_input\_tokens} to only pool information for it's assigned next span, we introduce a new argument to the interface --- \texttt{\textbf{encdec\_segment\_ids}}. 
In the original interface, \texttt{encoder\_segment\_ids} controls the attention mask of the encoder's self-attention pattern and the decoder's cross-attention pattern. The newly introduced argument separates this dual responsibility and therefore allows encoder self-attention and decoder cross-attention pattern to be customized independently. With this new change, the \texttt{encoder\_input\_tokens} can now attend to all tokens while \texttt{decoder\_input\_tokens} cannot.

\begin{itemize}
\item \texttt{encoder\_input\_tokens}: Tokenized input text.
\item \texttt{decoder\_input\_tokens}: A vector of \textbf{zeros(} \texttt{num\_text\_spans} \textbf{)}.
\item \texttt{decoder\_target\_tokens}: Class ids for each \texttt{decoder\_input\_tokens}.
\item \texttt{encoder\_segment\_ids}: Used for packing.
\item \texttt{decoder\_segment\_ids}: Used for packing.
\item \texttt{encoder\_positions}: Used for packing.
\item \texttt{decoder\_positions}: None.
\item \texttt{\textbf{encdec\_segment\_ids}}: Used for packing and controlling which encoded text span each \texttt{decoder\_input\_tokens} can attend to.
\end{itemize}

\section{Experiments}
We conduct our experiments on three tasks and compare Enc(M)T5 with (M)T5 and (m)BERT. All checkpoint used are publicly available. We hope the study can help readers decide which checkpoint to leverage when fine-tuning these tasks. For T5 checkpoints, we used the $1.1$ version\footnote{Changes made from T5 $1.0$: \url{https://github.com/google-research/text-to-text-transfer-transformer/blob/main/released_checkpoints.md\#t511}} because it does not mix the downstream task. Since there are many ways to convert these tasks to text-to-text, we provide a paragraph for each subsection describing or T5 baseline. 

For all our T5 and EncT5 experiments, we use and follow the fine-tuning setup in the T5X library\footnote{\url{https://github.com/google-research/t5x/blob/main/t5x/configs/runs/finetune.gin}}~\cite{roberts2022t5x} . For example, we use a batch size of $128$, $0.1$ dropout, train for 0.1M steps with $0.001$ learning rate with Adafactor. and report test set numbers (if available) with the checkpoint which achieved the best numbers from the validation set. We will describe task specific hyper-parameter choice such as sequence length in each subsection. One of the design choice is to make sure we are comparable with the BERT baseline. 

We conduct our experiments with TPU-v3 ranging from 16 to 64 chips depending on the model size. All our experiments can be done within a week and varies by how often we run evaluation.

\begin{table*}[!ht]
\caption{Results on the EURLEX57k test set. *Numbers extracted from Table 2. of the EURLEX57k paper ~\cite{chalkidis-etal-2019-large}. We select EncT5 and T5 checkpoints to run on testset with Micro-F1 on dev set. } 
\begin{adjustbox}{width=\textwidth}
\begin{tabular}{l|ccccccccc}
               & \multicolumn{3}{c}{ALL LABELS} & \multicolumn{2}{c}{FREQUENT} & \multicolumn{2}{c}{FEW} & \multicolumn{2}{c}{ZERO}     
                     \\  & RP@5 & nDCG@5 & Micro-F1 & RP@5 & nDCG@5 & RP@5 & nDCG@5 & RP@5 & nDCG@5 \\
                     \hline 
                BERT-base* & 79.6 & 82.3 & 73.2 & 83.5 & 84.6 & 68.6 & 63.6 & 2.8 & 2.3\\
                T5-base & - & - & 73.4 & - & - & - & - & - & -\\
                EncT5-base & 81.3&	83.8&	\textbf{75.5}&	84.9&	85.9&	73.6&	69.6&	1.1&	0.7\\
                \hline
            T5-large & - & - & 73.9 & - & - & - & - & - & - \\
            EncT5-large & 81.4 & 83.8 & \textbf{75.9} & 84.8 & 85.8 & 71.7 & 68.3 & 2.2 & 2.2  \\
            \hline
            T5-xl & - & - & 74.4 & - & - & - & - & - & - \\
            EncT5-xl & 81.1&	83.5&	\textbf{75.7}&	84.6&	85.7&	71.3&	67.9&	1.1&	0.4\\
            \hline
            T5-xxl & - & - & 75.0 & - & - & - & - & - & - \\
            EncT5-xxl & 80.9&	83.2&	\textbf{75.4}&	84.2&	85.2&	71.7&	68.2&	0.0&	0.0    
\end{tabular}
\end{adjustbox}
\label{tab:eurlex}
\end{table*}

\begin{table}[!ht]
\centering
\caption{Results on the UDPOS test set. *Numbers are from Table 2. of the XTREME paper ~\cite{hu2020xtreme}.}
\begin{adjustbox}{width=0.35\textwidth}
\begin{tabular}{l|c}
           \textit{Cross-lingual zero-shot transfer}            & Acc \\
                     
                     \hline
                     mBERT-base* & 71.5\\
                mT5-base & 71.6\\
                EncMT5-base & \textbf{78.0}\\
                \hline
            mT5-large & 71.7 \\
            EncMT5-large & \textbf{79.5}\\
            \hline
            mT5-xl & 74.6 \\
            EncMT5-xl & \textbf{81.0}\\
            \hline
            mT5-xxl & 73.4\\
            EncMT5-xxl & \textbf{80.1}
\end{tabular}
\end{adjustbox}
\label{tab:sp}
\end{table}

\subsection{GLUE}
\label{sec:glue}
We evaluate Algorithm~\ref{alg:cls} with GLUE, the General Language Understanding Evaluation benchmark is a collection of resources for training, evaluating, and analyzing natural language understanding systems \cite{wang2019glue}. We use an input sequence length (in tokens) of 512 and target sequence length of 84. Packing\footnote{Packing reduces padding by putting multiple sequence in one example and avoid interaction from different sequence through attention masks.}, a trick to improve accelerator utilization, is enabled for both T5 and EncT5. 

\paragraph{T5 baseline} We follow \citeauthor{2020t5} to convert GLUE tasks to text to text format. Details can be found in Appendix D of the T5 paper ~\cite{2020t5}.

We draw 3 conclusions on the effectiveness of Algorithm~\ref{alg:cls} from Table~\ref{tab:glue}. (1) In similar scale, base and large, both T5 and EncT5 outperforms BERT. (2) EncT5 is on par with T5 despite almost having half of the number of parameters as shown in Figure~\ref{fig:params_glue} and therefore EncT5 is more parameter efficient. (3) EncT5 also benefits from larger pre-trained T5's.  

\subsection{EURLEX57K}
We evaluate Algorithm~\ref{alg:mlabel} with EURLEX57K,
containing 57k English EU legislative documents
from the EUR-LEX portal, tagged with around 4.3k labels (concepts) from the European Vocabulary \cite{chalkidis-etal-2019-large}. We use an input sequence length (in tokens) of 512 for both T5 and EncT5. For target sequence length, we use 64 for T5 and 4271, which represents a latent vector for each label, for EncT5. 

\paragraph{T5 baseline} The inputs are the same for all models where we concatenate the header, the recitals, and the main body. For targets, we concatenate all labels with a comma and a single white space. We only report Micro-F1 because T5 does not provide a straightfoward way to get per token score. We will discuss more T5 drawbacks in Section 5.

A similar conclusion with Section~\ref{sec:glue} can be drawn from Table~\ref{tab:eurlex} with two distinction. (1) EncT5 outperforms both BERT and T5 in similar model scale. (2) Though T5's performance improves as we scale the model size, EncT5's performance does not seem to improve. It is unclear if this is a limitation of Algoithm~\ref{alg:mlabel} or the dataset since to our knowledge we did not observe a Micro-F1 higher than our reported numbers and we will leave as future work to conclude this.   

\subsection{UDPOS}
We evaluate Algorithm~\ref{alg:mlabel} with UDPOS, a Part Of Speech (POS) tagging data from the Universal Dependencies v2.5 (Nivre et al., 2018) treebanks, which cover
40 languages. Each word is assigned one of 17 universal
POS tags. We report the average accuracy of 33 languages. Similar to XTREME \cite{hu2020xtreme}, we adopt the zero-shot cross-lingual transfer scenario, where annotated training data is provided in English and evaluated on all 33 languages. mT5 ~\cite{xue-etal-2021-mt5} checkpoints are used for this task. The mT5 architecture and the training recipe closely follows  T5.1.1 but on the mC4 dataset.

\paragraph{T5 baseline} We mimic the span corruption task ~\cite{2020t5} to specify the input text spans in which the model should predict a Part of Speech tag on, and specify a sentinel token \texttt{<i>} after each text span. For targets, we specify the label after each corresponding sentinel token. For example in the example input text \texttt{cat eats fish}, the inputs are transformed to be \texttt{cat <0> eats <1> fish <2>} and the targets are \texttt{<0> NOUN <1> VERB <2> NOUN}. To calculate the metrics, we parse the targets and predictions with the following regular expression \texttt{<\textbackslash d+> } and only count the predicted label as correct if the sentinel token also matches.     

Despite moving to a multi-lingual setup, our conclusion is similar to the previous subsection. EncMT5 outperforms both mBERT when in similar model scale and outperforms mT5 in all model sizes. EncMT5 does seem to improve as we scale the model size but plateaus around xl/xxl size.

\begin{table}[!t]
\centering

\caption{Results on different EncT5 decoder layers and initialization where \texttt{12d}stands for 12 decoder layers.}
\begin{adjustbox}{width=0.4\textwidth}
\begin{tabular}{l|ccc}
               & GLUE & EURLEX & UDPOS    
                     \\ EncT5-base & Avg & Micro-F1 & Acc \\
                     \hline
                1d & \textbf{84.4} & 75.5 & 78.0 \\
                12d & 78.6 & \textbf{75.7} & \textbf{78.3} \\
                12d-T5-init & 79.2 & 72.7 & 74.5 \\
\end{tabular}
\end{adjustbox}
\label{tab:sp}
\end{table}

\section{Effectiveness of decoder in EncT5}
Despite being named EncT5, the framework does allow setting more than 1 decoder. The decoder weights can also be initialized from T5 weights other than from scratch. In this section, we study if these treatment results in better performance. We present the results in Table~\ref{tab:sp}. 

\subsection{Number of decoder layers}
We conducted a study to evaluate the effectiveness of decoder layers by using one or the same number of decoders as T5. We find that for GLUE, our classification experiment, the performance dropped. We further breakdown the summarized score in Table~\ref{tab:sp} and find that most of the drop comes from datasets containing only a few thousand training examples (such as \texttt{cola} or \texttt{rte}). This is reasonable as we now have more parameters to train. We also noticed that there is no improvement on large datasets such as \texttt{MNLI}. The reasoning for this is that the decoder does not play much of a role when there is only a single decoder token which makes self-attention an identity operation.  

In EURLEX and UDPOS experiments, we do see some gains in the cost of almost doubling the number of parameters. One explanation of the improvement is that for these two tasks, we have more than one decoder inputs which can benefit from multi-layers of attention. Given that the improvement is not significant and comes at the cost of more compute, we set the number of decoder layers in Algorithm ~\ref{alg:mlabel} and ~\ref{alg:tag} to 1 and retained the EncT5 name for the framework.

\subsection{Decoder layers initialization}
Because we are given a pre-trained T5 checkpoint with a trained 12-layer decoder, an obvious step was to try initializing both the encoder and decoder weights. However, loading the weights from T5 or mT5 was worse than random initialization. We hypothesize that this is due to the fact that the latent embeddings are recognized as special tokens such as \texttt{[BOS]} from the beginning for training.

\section{Why EncT5 over BERT?}

\subsection{Better performance}
Shown on three tasks, EncT5 is better than BERT when the models are in a similar scale. The reason behind this is a mixed of better pre-trained checkpoint and the EncT5 framework design. In GLUE, the benefit seems to come from the T5 checkpoint; in EURLEX and UDPOS, the EncT5 treatment seems more dominant. The finding suggests that users hoping to fine-tune pre-trained weights to be used for classification, multi-label, structured prediction, should go for EncT5.  

\subsection{Large scale checkpoint availability}
Architectures capable of serving the use cases of a larger variety of discriminative and generative tasks, such as the encoder-decoder transformer, have become increasingly popular and with more model checkpoints at increasing parameter sizes readily available for the public. Because of the increasing compute and energy cost to pre-train large-scale models, it is increasingly costly to regenerate encoder-only models and thus they have become less available. Therefore, it is important to find ways to reuse and repurpose existing pre-trained encoder-decoder models to conserve compute costs while still optimizing for discriminative tasks.

\section{Why EncT5 over T5?}
\subsection{Efficiency}
EncT5 is more efficient than T5 in two ways: (1) combining decoder inference in one single step increases parallelism, and (2) pruning parameters.  Pruning the decoder layers improves efficiency in both training and inference time as well as number of parameters.   

\subsection{Additional Model Outputs (Usability)}
In some cases, the user not only expects the prediction but also the score. For example, in multi-label classification, it is common to tune the binary decision threshold as opposed to using $0.5$ to adjust the decision boundary for business reasons. While it is possible for T5 to extract per-token score by calculating the likelihood of the inputs and a label, additional inference compute is needed. In the case of EURLEX57K where there is 4.3k labels, the user would need to run inference 4.3k times. In EncT5, the model can return the score and prediction together without additional compute cost.

\subsection{Set Prediction (Quality)}

The multi-label problem is essentially a set prediction problem where the order of the labels do not matter. However, the language model in T5 might learn from the order we provided. For instance, if $label_j$ always come after $label_i$ in the training set, it is very hard for the model to predict $label_j$ if $label_i$ is not generated. One solution is to provide all permutations when converting the labels to a string during training and ensemble the beam search results at inference. However, this would increase computation significantly.

\subsection{Constrained Decoding Output (Quality)}

In T5, the auto-regressive model determines when it is done emitting output and users are left to parse and transform the unstructured string output to their structured expectations. T5 does not have any constraint on the decoding algorithm, so there is no guarantee that the output follows an expected structure and may thereby hindering the output's usability. We believe this is the main reason why EncT5 outperforms T5. 

The T5 outputs from our UDPOS experiment in Section 4.3 encountered this shortcoming. For example, the model generated the prediction \texttt{<0> ADJ <1> PART <2> VERB <3> PUNCT ... <26>} which ends with 26 sentinel token while the target has 56 sentinel tokens. In EncT5, it is possible to encode the desired output structure and request 56 predictions by passing 56 decoder input tokens.

\section{Related Work}
Recent NLP progress can be attributed to un-supervised pre-training and scaling. Encoder-only~\cite{devlin-etal-2019-bert, roberts2022t5x, yang2019xlnet} and Encoder-decoder~\cite{lewis-etal-2020-bart, 2020t5} on one hand improves the transfer-learning scenario where the model is pre-trained and then fine-tuned with the downstream labeled task. Decoder-only models on the other hand improves the few-shot learning scenario where models are first pre-trained and then prompts~\cite{brown2020language} are used to provide instructions or examples of the downstream task. Our work falls in the transfer learning scenario where we fine-tune pre-trained checkpoints with downstream tasks. 

The boundaries of transfer-learning has been pushed by scaling both model and data size through techniques like model parallelism~\cite{shazeer2018mesh}. The resources required scales with model size for unsupervised pre-traing scales. Thus, architecture that are capable of being used for many down-stream tasks are in favor. The introduction of decoders in T5 unlocks the ability to fine-tune generation tasks. However, in this work we show that T5 does not have to be used in an auto-regressive fashion through the following common Transformer~\cite{vaswani2017attention} tricks: (1) leveraging attention masks to control the access of the context to each token~\cite{dong2019unified}; and (2) feeding a latent vector to the Transformer to learn how information should be aggregated~\cite{pmlr-v139-jaegle21a, carion2020end}.         

The closest to our work from a modeling perspective is the sentence representation work~\cite{montero2021sentence}, which pools the encoder tokens with a multi-head attention as the auto-encoder bottleneck while keeping the encoder fixed.  

\section{Conclusion and Future Work}\label{sec:conclusion}
In this work, we propose a framework, EncT5, which converts an encoder-decoder T5 model to a non-auto-regressive model for classification, multi-label, and structured prediction and results in better performance, efficiency and usability. We also show that with models of sizes in similar ballpark, EncT5 is a better choice than BERT. However, we have not yet concluded that Encoder-decoder pre-training is better than Encoder-only pre-training when everything else are hold the same (dataset, number of tokens seen). We did attempt to pre-train BERT on C4 but realized that designing an apple to apple comparison is a topic of its own. 

For future work, there are two directions: (1) the encoder-only model direction; and (2) the decoder-only model direction. For (1), we would like to answer the question of whether the EncT5 gains over BERT comes from pre-training or fine-tuning. We can do this by also training BERT on C4 with a similar training schedule. For (2), as decoder-only model becomes prevalent ~\cite{GPT3_NEURIPS2020, chowdhery2022palm}, we would also like to study how to convert them to a non-auto-regressive one to tackle the tasks we listed in the paper. These tasks covers most of the use cases in NLP application and is worth investigating if auto-regressive decoding is really necessary.   


\section{Limitations}
Our work builds on top of publicly available pre-trained checkpoints and study the fine-tuning performance on top of these checkpoints. BERT and T5 are trained on different datasets with different objectives. Our work does not answer whether C4 is better than Wikipedia+Book Corpus nor whether span corruption (Encoder-Decoder) is better than next sentence prediction + masked language model (Encoder-only). 

\bibliography{anthology,custom}
\bibliographystyle{acl_natbib}

\appendix

\end{document}